# Human Attention Estimation for Natural Images: An Automatic Gaze Refinement Approach

Jinsoo Choi, Tae-Hyun Oh, *Student Members, IEEE,* and In So Kweon, *Member, IEEE*

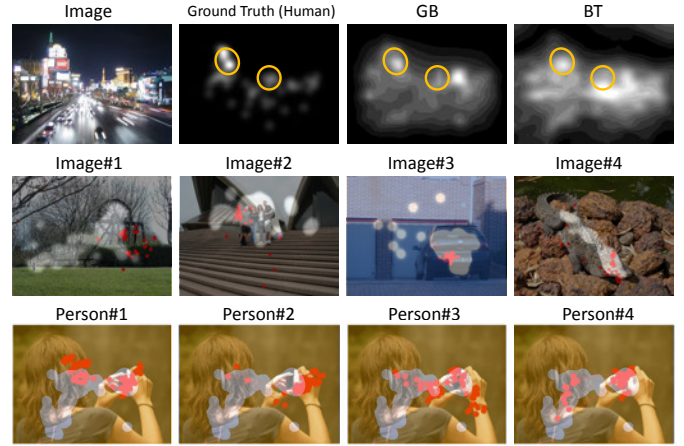

Fig. 1: **Illustration of key observations.** [Top] Computational saliency models show high false positives on natural scene photos, but also show true positive responses. [Middle] Implicitly calibrated gaze estimation with a simple configuration shows consistent estimates despite marginal errors. [Bottom] Although people share common viewing traits, gaze patterns differ by individuals. Images are from Judd *et al.* [27]. The ground truth fixation regions are overlaid on images while red dots indicate gaze prediction positions. Please refer to color version.

*Abstract*—Photo collections and its applications today attempt to reflect user interactions in various forms. Moreover, photo collections aim to capture the users' intention with minimum effort through applications capturing user intentions. Human interest regions in an image carry powerful information about the user's behavior and can be used in many photo applications. Research on human visual attention has been conducted in the form of gaze tracking and computational saliency models in the computer vision community, and has shown considerable progress. This paper presents an integration between implicit gaze estimation and computational saliency model to effectively estimate human attention regions in images on the fly. Furthermore, our method estimates human attention via implicit calibration and incremental model updating without any active participation from the user. We also present extensive analysis and possible applications for personal photo collections.

## I. Introduction

People can access abundant sources of photo collections available in social networks and cloud services (*e.g.*, Facebook, Instagram, Flickr, iPhoto, Picasa) today. Nowadays, users can use automatic photo applications embedded in photo collection services (*e.g.*, face tagging, tag region suggestion), which allow various forms of interaction. In fact, photo applications today attempt to produce interactive results reflecting human intentions by exploiting human behavior traits shown while accessing photo collections.

Due to similar motivations recently, computational modeling of human visual attention has attracted the interest of analysts and psychologists as well as computer vision researchers. Especially, attention modeling on visual data has been researched as a form of memorability [24], aesthetics [14], interestingness [19] or saliency [27]. Since human attention regions can be obtained by automatically quantifying subtle insights from human visual behavior, these models have been attempted in various fields including computer vision applications [39], [30], [49], [33], [45], [10], and also non-engineering fields such as marketing and advertisement [43]. However, predicting interestingness (or ROI) is still challenging only with learned or designed computational models, because it is highly subjective and dependent on specific categories [19].

Hence, some personal photo collection applications involve explicit user interactions [31], [51] such as active sketching, dragging of photo components, tagging or labeling to reflect the users' subjective attention preferences and produce user oriented results. Apart from other interaction based methods, we pursue a simple method that captures individual user preferences by gaze estimation without involving the users' notice or active participation, but with an off-the-shelf device, *i.e.*, a web-cam. The key idea is to integrate rough gaze estimates and a computational attention model (in this paper, we use a computational saliency model). As illustrated in Fig. 1, the key observations are as follows. Firstly, we observe that, in spite of the difficulty to accurately predict salient regions in natural scenes, modern saliency models show reasonable agreement with salient human attention regions by virtue of the advances on computational saliency research (See Fig. 1-[Top]). Notice the saliency maps show responses to local image features leading to extensive false positive responses, but also show appropriate responses at true attention locations obtained by an accurate eye tracking device. Secondly, implicitly calibrated gaze estimates via simple configuration show rough but plausible consistency with expensive eye tracker results (ground truth) (See Fig. 1-[Middle]). Lastly, the gaze estimates of individuals show different fixation behaviors, while they also reflect common viewing traits (See Fig. 1-[Bottom]).

Based on the observations, we develop a fully incremental and simple human attention map estimation method by integrating existing saliency models and implicit gaze estimation

Jinsoo Choi, Tae-Hyun Oh and In So Kweon (corresponding author) are with the School of Electrical Engineering, Korea Advanced Institute of Science and Technology (KAIST), Republic of Korea. E-mail: jschoi@rcv.kaist.ac.kr, thoh.kaist.ac.kr@gmail.com, iskweon@kaist.ac.kr

method. This integration allows our method to reflect personal attention preferences which can be used in a number of personal photo applications. Moreover, our method does not require any active participation. In a task as challenging as accurate human attention estimation on natural scene images, we utilize a behavior humans naturally do when encountered by a photo: look at the photo *i.e.*, gaze behavior. In short, our human-computer method does not require a training step and the user to actively execute a specified interaction guideline. The only thing a user has to do in our framework is to just view photos in a photo collection (which is a highly natural process as mentioned).

The preliminary version of this paper has appeared in [3]. We extend [3] by addressing our motivation along with detailed explanation on gaze estimation via implicit calibration in Sec. III-A and the refinement approach in Sec. III-B. The experimental setup and analysis has been entirely revised in Sec. IV, and we show a number of applications in Sec. V.

Overall, our contributions can be summarized as follows:

- We develop a human visual attention estimation method on the fly for natural image collections without any special hardware settings, based on implicit calibration and incremental model updating. While users use photo collection services, our proposed method can learn user-specific parameters without the user noticing.
- We experiment systematically to show the validity and flexibility of the proposed method with 10 subjects.
- To demonstrate how the proposed method can be utilized in practice, we show three possible applications for personal photo collections: saliency map refinement, tag candidate region suggestion, and photo retrieval by personal interest location query.
- If this paper is accepted, we will publish our dataset collected on top of MIT1003 [27] (which is used for our experiments). The dataset includes information no other gaze datasets provide, consisting of actual human eye images, gaze estimation data obtained from implicit calibration of 10 individuals.

## II. RELATED WORK

**Gaze estimation** Gaze estimation is the process of estimating where a user is looking at, which can be used for constructing an attentive human-computer interaction interface [22], [47]. Gaze estimation methods are boiled down to either model-based or appearance-based methods[1]. Model-based methods [20] typically involve specialized hardware and explicit 3D geometric eye modeling which require sophisticated and extensive calibration. Most of the time, model-based approaches require the person to wear a head mount close up IR camera device whereas the appearance-based method simply involves a single camera setting anywhere in the PC environment. We do not want to set up specialized acquisition settings since people usually do not set up IR cameras or specialized configurations around their PC environment just for photo application purposes. We carry out an appearance-based method with an off-the-shelf web-cam which is commonly set up or built-in in PCs nowadays.

Works addressing gaze estimation by utilizing saliencies include Sugano *et al.* [45] and Chen *et al.*[12]. While Chen *et al.*achieve relatively higher accuracy, they address a model-based setup with special camera settings and require a long data capture time. Sugano *et al.*construct a gaze estimator by using eye images captured from a user watching a few video clips. Video based approaches, including Sugano *et al.*, Feng *et al.* [16] and Katti *et al.*[28], benefit from temporal cues in the form of temporal saliency aggregation or tracking. Our work involves independent photos instead of videos, thus we cannot benefit from reliable saliency supports with inter-frame temporal relationship. Despite this, by aggregating independent saliency maps via incremental soft-clustering, we can enhance the gaze estimation quality even with a few independent images and can reduce data redundancy. As for appearance-based gaze estimation involving independent images, Alnajar *et al.*[1] propose a data driven approach, where a rough input gaze pattern is transformed to the most similar template pattern in a predefined database. This method requires numerous human fixation data with high accuracy.

**Computational saliency models**   Predicting saliency for natural scene photos still remains as a challenging task. Over the years, visual saliency researches have aimed to computationally model and predict human attention behavior given visual data. Many attention models have been developed under a pioneer work, Feature Integration Theory (FIT) [46], that deals with which visual features are important and how they should be combined to reveal conspicuity. Biologically inspired by the center-surround structure of receptive fields, Itti *et al.* [25] was the first to computationally search patterns from multiple features on top of FIT. Following this pioneering work, most of the saliency models [21], [26], [18], [4], [9], [41], [27], [23], [17], [32], [5], [50], [40] are integration approaches that compute saliencies by combining multiple visual features extracted from images according to an attention model criteria: rarity [5], spectral analysis [23], graph model [21], information theoretic model [9], *etc.*. For thorough reviews in saliency estimation, one can refer to Borji *et al.* [6], [7], [8] and Riche *et al.* [38].

Apart from biologically inspired models or heuristic feature integrating methods, there have been machine learning approaches which learn the relationship between visual features and human attention behavior. Judd *et al.* [27] and Kienzle *et al.* [29] learn a SVM from low, mid, high-level features and human fixation data to construct a saliency model. Recently, deep neural networks by Shen *et al.* [42] and Vig *et al.* [48] have shown relatively improved results in natural scenes due to its ability to represent complex models without explicitly learning high-level context models like car or face detectors.

Our motivation is supported by Judd *et al.*[27] and Alnajar *et al.*[1], where they reflect the fact that humans show similar viewing patterns when they look at a stimulus (we call *commonness*). In general, computational saliency models represent this commonness and are used extensively in our work. In our approach, saliency maps are used for inferring gaze positions in the gaze estimation step. Also, our method integrates the roughly estimated gaze positions with a computational saliency baseline and uses it for each independent image as a guide to obtain a refined attention map. By the integration, our method can effectively estimate person-specific attention regions in natural image collections without any special settings or tedious processes.

---

[1]Thorough reviews on camera based gaze estimation methods including commercial products can be found in Hansen *et al.*[20]

### III. AUTOMATIC GAZE REFINEMENT APPROACH

In order to estimate person-specific attention on natural scene images without special settings, we undergo two stages namely: (1) initial gaze estimation via implicit calibration and (2) integrating initial gaze estimation and computational saliency maps.

#### A. Initial gaze estimation via implicit calibration

In this stage, our goal is to construct a gaze estimator via implicit calibration. Specifically, our algorithm learns the user's gaze behavior while the user sees photos in a collection. From the photos, let's consider a person sees a few images denoted by $\{I_1, ..., I_N\}$, where $N$ denotes the number of images and $I_n \in \mathbb{R}^{p \times q}$ where $p \times q$ denotes the image dimensions. While the person is viewing an image, a single camera can capture the viewer's face. We capture $M$ frames[2] of the person's frontal view per image seen, a total of $NM$ frames of the person's face is captured. We extract the eye images by utilizing Active Shape Model (ASM) [35] and use its gray scale eye images as the vectorized form $e_{nm} \in \mathbb{R}^d$, where $n$ and $m$ denote the indexes of the $n$-th image and the $m$-th eye image respectively, and $d$ denotes the eye feature vector dimension.

In order to regress the relationship between eye images and fixation location, *i.e.*, a gaze estimator, we need not only eye feature data but also the corresponding fixation positions. Since the user does not undergo an explicit calibration scheme for gaze training and views images instead, we regard the saliency maps of the viewed images as fixation position probability distributions based on our observations discussed in Sec. I and II. We apply a computational saliency model combining low-level features such as orientation, color, intensity, bottom-up features with top-down cognitive visual features [4]. However, a single saliency map per se may not be a reliable fixation position indicator. Imagine that a user naturally looked at the same position on different images and we aggregate those image saliency maps. Then, it is highly likely to produce a vivid peak of saliency around that position in the aggregated map due to the consensus of independent saliency maps. Moreover, since the saliency map [4][3] is computed by a combination of low and high-level features, it is expected to produce high true positive responses (*i.e.*, salient regions near human fixation positions) and thus produce a reliable fixation position indicator when aggregated.

**Aggregation stage** We cluster the collection of eye features by Gaussian Mixture Model (GMM) [2] to reduce redundant data and effectively train the gaze estimator. Simultaneously, it allows us to estimate weights for saliency aggregation with respect to the clustered eye features. Each Gaussian component is represented as:

$$P(\boldsymbol{e}_{nm}|z_{nm} = c) = \mathcal{N}(\boldsymbol{e}_{nm}|\boldsymbol{\mu}_c, \boldsymbol{\Sigma}_c), \quad (1)$$

where $z_{nm}$ denotes a latent variable of $\boldsymbol{e}_{nm}$ for the membership of the cluster, $\boldsymbol{\mu}_c$ and $\boldsymbol{\Sigma}_c$ are the mean and covariance of the $c$-th cluster respectively. Since the most computationally expensive part of our algorithm is this eye feature clustering, batch GMM [2] is used only at the initial stage, while we use incremental GMM [11] for later updates[4] to reduce computations. We set the number of clusters $K$ to 30 throughout our entire experimental procedure.

Aggregation of the saliency maps is done by taking the weights given to eye feature vectors by GMM and simply applying them to the corresponding collection of saliency maps. Specifically, for the training images $\{I_n\}$, we compute the corresponding normalized saliency map $\{s_n\}$, where $s_n \in [0, 1]^{p \times q}$. Saliency maps are aggregated with weights from eye feature membership probability as:

$$\overline{\boldsymbol{p}}_c = \frac{\sum_{n,m} P(\boldsymbol{e}_{nm}|z_{nm} = c)\boldsymbol{s}_n}{\sum_{n,m} P(\boldsymbol{e}_{nm}|z_{nm} = c)}, \quad (2)$$

producing a fixation position probability map $\overline{\boldsymbol{p}}_c \in \mathbb{R}^{p \times q}$. These will produce reliable pairs of fixation position probability maps and eye feature cluster centers by suppressing irrelevant saliency effects.

**Learning the relationship between eye feature and implicit gaze** To predict the gaze positions, we learn Gaussian process regression (GPR) [37] between eye feature cluster centers and corresponding gaze positions, *i.e.*, , we construct an estimator that predicts the distribution $P(\boldsymbol{g}^*|\boldsymbol{e}^*, \mathcal{D}_g)$ given a new eye image $\boldsymbol{e}^*$ and the training data $\mathcal{D}_g = \{(\boldsymbol{\mu}_1, \boldsymbol{g}_1), ..., (\boldsymbol{\mu}_K, \boldsymbol{g}_K)\}$, where $\boldsymbol{g}_c \in \mathbb{R}^2$ denotes the 2D gaze position corresponding to the eye feature cluster center $\boldsymbol{\mu}_c \in \mathbb{R}^d$. However, we only have a dataset pairing eye feature cluster centers and fixation probability maps, *i.e.*, the aggregated saliency maps, as $\mathcal{D}_p = \{(\boldsymbol{\mu}_1, \overline{\boldsymbol{p}}_1), ..., (\boldsymbol{\mu}_K, \overline{\boldsymbol{p}}_K)\}$, instead of the set $\mathcal{D}_g$ of eye features and gaze points. By simply peaking the maximum probability location of $\overline{\boldsymbol{p}}_c$ as $\boldsymbol{g}_c$, we construct the training set $\mathcal{D}_g$.

We independently model and regress $x$ and $y$ components of gaze $\boldsymbol{g} = [g_x, g_y]$, so that we reduce the problem dimension and achieve efficiency. For simplicity, we describe the $x$ component $g_x$ only, but deriving $g_y$ is analogous. We assume a noisy observation model as $g_x^* = f(\boldsymbol{e}^*) + \epsilon$ with noise $\epsilon \sim \mathcal{N}(0, \gamma)$, where $\gamma \geq 0$. GPR assumes that data can be represented as a sample drawn from a Gaussian distribution, thus $f(\cdot)$ is modeled by GPR with zero-mean multivariate Gaussian distribution as

$$P(g_x^*, \boldsymbol{e}^*, \mathcal{D}_g) = \left[\begin{array}{c} \overrightarrow{\boldsymbol{g}}_x \\ g_x^* \end{array}\right] = \mathcal{N}\left(\boldsymbol{0}, \left[\begin{array}{cc} \mathbf{K} & \mathbf{K}_*^\top \\ \mathbf{K}_* & \mathbf{K}_{**} \end{array}\right]\right), \quad (3)$$

---

[2] In this paper, we use the fixed number $M$ for simplicity of the representation. In practice, it can be arbitrary values.

[3] We mainly use the method suggested by Borji *et al.*[4], but any method can also be used in our framework. We provide the effects of our algorithm with respect to multiple baseline saliency methods in Sec. V.

[4] Details on update rules for eye clustering are derived in the appendix.

**Algorithm 1** Training phase - Implicit gaze calibration
1: display $I_n$
2: capture and save eye features $e_{n,\{1:M\}}$
3: **if** $n = b$ (batch size) **then**
4:    batch GMM clustering of eye features
5: **else if** $n > b$ **then**
6:    incremental GMM clustering of eye features
7: **end if**
8: **for** $j = 1 : K$ **do**
9:    $\overline{p}_c = \frac{\sum_{n,m} P(e_{nm}|z_{nm}=c) s_n}{\sum_{n,m} P(e_{nm}|z_{nm}=c)}$
10:    take maximum and construct $\mathcal{D}_g$
11: **end for**
12: Train GPR with $\mathcal{D}_g$

**Algorithm 2** Testing phase - Integration stage
1: Output gaze estimation results from GPR
2: **for** $n = 1 : N$ **do**
3:    mean shift (initialize at gaze points)
4:    convolve a Gaussian kernel
5: **end for**
6: Output attention maps

where $\mathbf{K} \in \mathbb{R}^{K \times K}$ and $\mathbf{K}_{(i,j)} = k(\boldsymbol{\mu}_i, \boldsymbol{\mu}_j) + \gamma \delta_{ij}$,
$\mathbf{K}_* \in \mathbb{R}^{1 \times K}$ and $\mathbf{K}_{*(j)} = k(e^*, \boldsymbol{\mu}_j)$,
$\mathbf{K}_{**} = k(e^*, e^*) \in \mathbb{R}^{1 \times 1}$,

$\overrightarrow{\mathbf{g}}_x = [g_1^x; \cdots; g_K^x]$ from $\mathcal{D}_g$, $k(\cdot, \cdot)$ denotes a kernel (or called as covariance function), and $\delta_{ij}$ is the Kronecker delta which is 1 if $i=j$ and 0 otherwise. We adopt the $d$-th order polynomial kernel defined as

$$k(\boldsymbol{u}, \boldsymbol{v}) = \alpha (\beta + \boldsymbol{u}^\top \boldsymbol{v})^d, \quad (4)$$

where $\alpha > 0$ and $\beta \geq 0$. We use the 3-rd order polynomial, *i.e.*, $d = 3$. Then, the hyper-parameters to be learned are $\alpha, \beta, \gamma$ in our model. We learn them from the training data $\mathcal{D}_g$ by GPML toolbox [36]. Note that, since we use the clustered data $\{\boldsymbol{\mu}_c\}$, the computational complexity in the learning phase mainly depends on the number of clusters $K$ and the dimension of eye feature $d$. Hence, the hyper-parameter learning is virtually fast with the small number $K$ and the down-sampled eye feature vectors.

In contrast to the squared exponential (SE) kernel[5] typically used in GPR, we use the polynomial kernel. It is known that the polynomial kernel leads to the function shape rapidly decreasing and preserves high frequency regression, while the SE kernel affects in a broader range due to the heavier tail than that of polynomial and behaves like a low-pass filter which smooths out the solution space [44]. Since the eye images have similar appearances and the gaze estimation is required to be accurate, the polynomial kernel would be a better choice.

After learning the hyper-parameters of GPR, we can infer the gaze estimate. From the joint distribution in Eq. (3), what we are interested in is the conditional probability $P(g_x^* | e^*, \mathcal{D}_g)$ to predict the gaze component $g_x^*$ for a newly observed eye image $e^*$. Since the conditional probability also follow Gaussian, we can compute the posterior mean by the closed-form solution:

$$g_x^* = \mathbf{K}^* \mathbf{K}^{-1} \overrightarrow{\mathbf{g}}_x. \quad (5)$$

Note that, in the above equation, $\mathbf{K}^{-1} \overrightarrow{\mathbf{g}}_x$ can be computed in advance in the learning phase, and predicting $g_x^*$ can be done by cheap vector multiplications. A pseudocode of this stage is shown in Alg. 1.

[5]$k(\boldsymbol{u}, \boldsymbol{v}) = \alpha \exp(-\beta \|\boldsymbol{u} - \boldsymbol{v}\|^2)$.

### B. Integrating initial gaze estimation and computational saliency map

The basic notion of this stage is to estimate human attention regions on each image by comparing and contrasting the initial gaze estimation result and the computational saliency map of the image. Typically, gaze estimation via implicit calibration inevitably has lower accuracy compared to the explicit calibration scheme. This is due to the fact that explicit calibration involves ground truth fixation points, whereas implicit calibration involves deduced gaze points from probability maps instead. For this reason, the initial gaze estimation results from the previous stage alone is not reliable for image attention region estimation. Similarly, although many computational saliency models produce high overlap with actual human fixation positions, they also retain high false positive regions. Natural scenes have unimaginable variety in content and subtle characteristics that make it difficult for current state-of-the-art saliency models to predict salient human attention regions accurately. Thus, computational saliency maps alone are not reliable either for our purpose.

Despite these limitations, initial gaze estimations and computational saliency maps provide valuable information and cues. Our goal is to utilize the initial gaze estimation as a rough indicator of attention regions, and potential true positive responses on saliency maps as a useful prior of human attention candidates for each viewing image. We integrate them to detect true salient regions produced from the computational saliency model and produce a reliable attention estimation via a mode seeking algorithm.

The basic idea is to seek out salient regions around the initially estimated gaze positions through greedy search which is practically done by a mean shift algorithm [13]. The mean shift algorithm does not require prior knowledge on the number nor shape of distribution. While the mean shift algorithm traditionally finds a mode from samples drawn from an arbitrary mother distribution by kernel density estimation, in our problem, the saliency map $\hat{s}$ of the viewed image $\hat{I}$ can be regarded as a 2D density map directly. Since we collect $M$ eye feature vectors for a viewed image $\hat{I}$, we have the initial gaze estimates denoted as $\{\hat{g}_1^{(0)}, ..., \hat{g}_M^{(0)}\}$. These gaze estimates mark where to start the mode seeking algorithm. Then, the problem is simplified to:

$$\hat{g}^{(k+1)} = \arg\max_{x \in \mathcal{N}_h(\hat{g}^{(k)})} \hat{s}(x), \quad (6)$$

where $\hat{g}^{(k)}$ denotes a gaze point at the $k$-th iteration, $\mathcal{N}_h(x)$ is the neighbor region at $x$ within the window size $h$. We iteratively solve Eq. (6) for each gaze point by greedily shifting to a maximum value within the window size at every iteration.

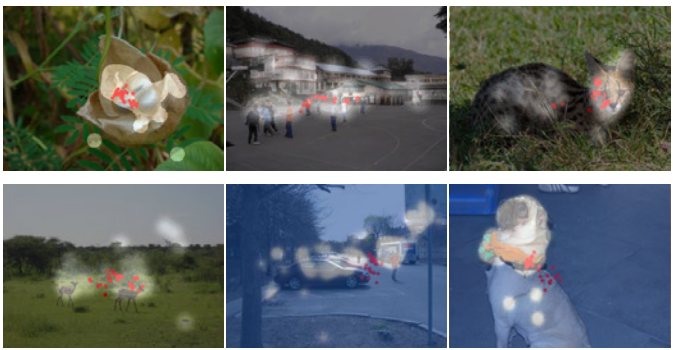

Fig. 2: **Initial gaze estimation.** This figure shows (Top) accurate gaze estimation examples (red dots) on natural scene images and (Bottom) misaligned gaze estimation examples. The ground truth fixation regions are overlaid on images. Please refer to color version.

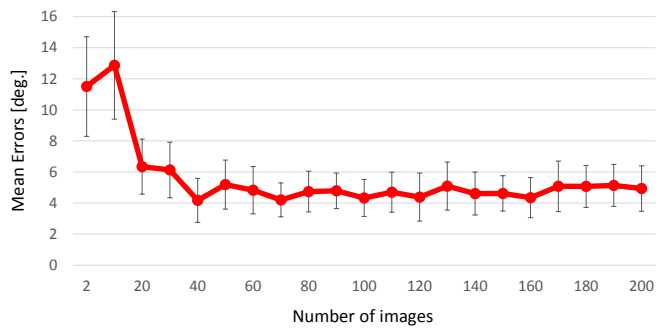

Fig. 3: **Mean error vs. number of images seen.** As the user views images, online GMM incrementally clusters data for regression. The graph shows mean error (in degrees) with respect to the number of images seen.

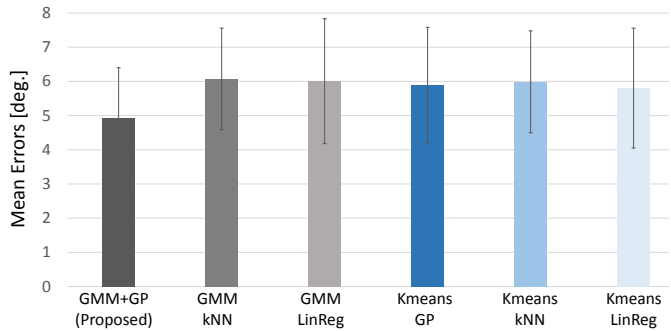

Fig. 4: **Gaze estimation method comparison.** This shows the mean estimation errors of the proposed method and other methods combining k-means clustering, k-nearest neighbor and linear regression.

We terminate the algorithm once the movement is converged. Since our method is inspired by mean shift, convergence to a local maximum is guaranteed [13] and therefore guarantees convergence towards a salient region nearby the estimated gaze positions. These resulting $M$ convergence positions on an image saliency map are convolved with a Gaussian kernel to produce an estimated attention map. A pseudocode of this stage is shown in Alg. 2.

## IV. EXPERIMENTS ON GAZE AND ATTENTION REGIONS

In this section, we describe how we conducted experiments for implicit gaze estimation method and attention region generation. The experimental procedure was divided into two parts corresponding to the initial gaze estimation step and integration step. We evaluate on data obtained from 10 subjects viewing photos selected at random.

Computational time was measure for batch GMM of 20 images (0.8-2.5s), incremental GMM of 5 images (0.5s), saliency aggregation (0.45s), GPR training (0.1-0.5s), and GPR testing (0.005s) on MATLAB.

### A. Initial gaze estimation evaluation

A subject is allowed to view a total of 200 natural scene photos on a flat screen display (1-1.5 seconds each) just the way a person will do when viewing images on a PC. During every 1-1.5 seconds of image display, the web-cam captures 30 frames of the subject's frontal view. Eye images are extracted via ASM [35] and their gray scale images resized to $3 \times 5$ are used as eye feature vectors for efficiency. The subjects are approximately 60 cm from the screen and the images are displayed in $1200 \times 1600$ resolution. We use natural scene images from the MIT1003 [27] dataset and the display images are selected at random for every subject. Since our method addresses implicit calibration using independent images, we choose to aggregate independent saliency maps with the incremental soft-clustering method. Fig. 2 shows some examples of gaze estimation results on images. Although gaze results provide rough attention region estimates, one can observe that some estimation error is present.

First of all we investigate gaze estimation accuracy depending on the number of images seen. Fig. 3 shows gaze estimation error with respect to the number of images seen. Estimation error is taken as the mean error in degrees on a test data. The test data image contains predefined points on which the subjects are instructed to fixate. A total of 450 test images with 9 predefined points are used to measure mean estimation error in degrees. As shown in Fig. 3, as the number of images seen increases, the gaze estimation error falls leading to higher accuracy. Notice that even by viewing 20 images, it gives reasonable accuracy of around 6 degrees error. From 40 images, we see the graph already hitting plateau phase. This signifies that people only need to look at around 20 to 40 images to obtain reasonable gaze estimation quality.

For quality assessment of our algorithm, we measure the gaze estimation error in comparison with a combination of different clustering and regression algorithms including k-means clustering which is used in Sugano *et al.* [45], k-nearest neighbor and linear regression. Fig. 4 shows that our method produces lower error than other baseline methods.

### B. Evaluation on attention maps

Here we evaluate our integration algorithm. To assess the quality of the attention maps, we utilize the ground truth fixation data provided in the MIT1003 [27] dataset. The fixation data provides all fixation positions of 15 subjects acquired from an accurate eye tracking device. The union of these fixation regions reflect virtually all possible human attention regions on an image, and we take this as ground truth fixation regions. Since the ground truth fixation data contains results from multiple subjects, it represents a generalized fixation map. On the other hand, our result attention maps are specified to each subject reflecting individual attention maps (since these are to be used in applications capturing user intentions). Thus we measure the Szymkiewicz-Simpson coefficient (SSC) which measures the overlap coefficient between our individual attention maps and ground truth fixation maps as defined as

$$OV(A, F) = \frac{\mid A \cap F \mid}{\min(\mid A \mid, \mid F \mid)} \quad (7)$$

where $|A|$ denotes the number of non-zero components of $A$, $A \in \mathbb{R}^{1200 \times 1600}$ is the attention map estimated by our algorithm and $F \in \mathbb{R}^{1200 \times 1600}$ is the ground truth fixation region map.

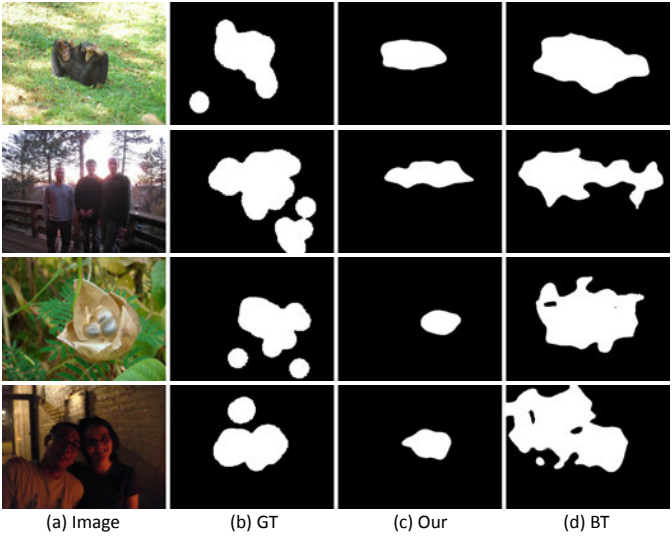

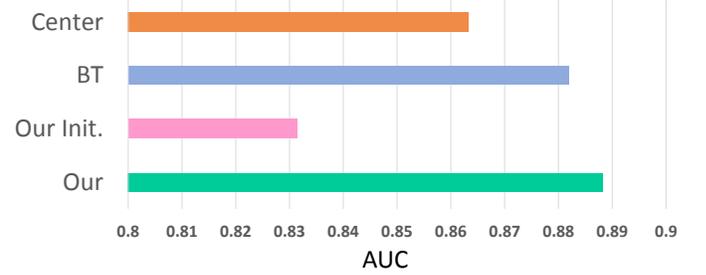

Fig. 7: **Evaluation on attention map by AUC [27].** We compare the AUC values with center bias, BT [4], our initial gaze, and our attention estimation map.

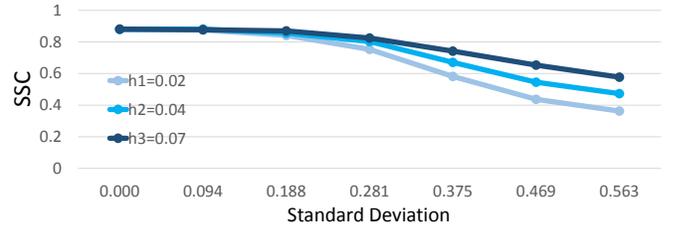

Fig. 8: **Effect of noise on mode seeking.** We show SSC values on attention maps produced from mode seeking with three different window sizes at varying noise levels. This was done in normalized coordinates.

Fig. 5: **Attention region comparison.** This figure shows (a) natural image stimulus, (b) thresholded ground truth fixation region maps, (c) our attention region estimations and (d) BT [4] computational saliency map.

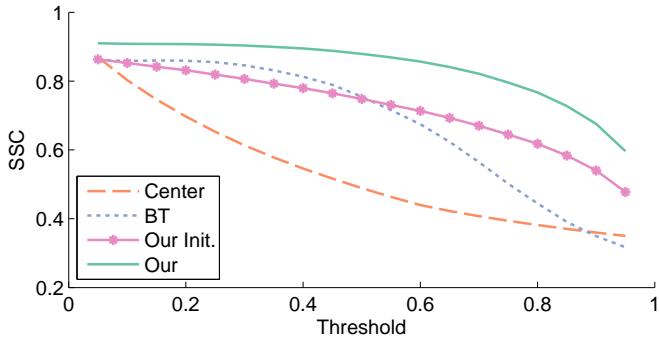

Fig. 6: **Evaluation on attention map by Szymkiewicz-Simpson coefficient (SSC).** We compare the coefficient values with a center bias map, BT [4] computational saliency map, our initial gaze map, and our attention estimation map. The horizontal axis denotes the percentage threshold, the vertical the score.

Since our method deals with attention maps of individual subjects, the quality assessment method differs from that of Judd *et al.*[27] which deals with assessing the *general* quality.

We measure the SSC to observe how much each individual attention map actually contain true fixation components via measuring overlap with the fixation map containing virtually all fixation regions. The coefficient thus can reflect how reliable our individual attention maps obtained from a cheap setting is compared to the ground truth fixation data collected from an accurate eye tracking device.

If the coefficient is close to 1, it signifies that the attention map reflects true positives and its response has high overlap with the ground truth fixation map. If the coefficient is close to 0, the estimated attention map has high false positives and low overlap with the ground truth map.

To evaluate with the SSC measure, we threshold our attention region estimation map via varying thresholds. In Fig. 5, we show example comparisons with BT [4] saliency. It shows that our attention maps have greater overlap with ground truth maps while BT saliency reflects false positives. In addition, we also compare our results with a center bias map. Fig. 6 shows the comparison with our results, BT saliency, center bias map, and our initial gaze estimation results with varying threshold percentages. Our results show higher values for all thresholds. The same thresholding scheme from [34] was applied. We also show results using the AUC metric in Fig. 7.

The integration stage involves using initial gaze estimates as mode seeking starting points on BT [4] saliency map. To evaluate the robustness and reliability of the mode seeking method, we apply Gaussian noise to the estimated gaze positions and measure SSCs on resulting attention maps as shown in Fig. 8.

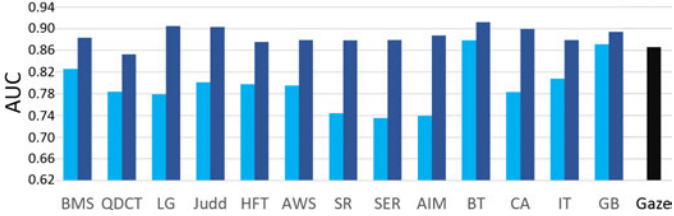

Fig. 9: AUC values are shown for baseline methods (light blue) and our method (dark blue). Average AUC was taken on 40 evaluation photos. Our initial gaze result is shown in black.

|  | Before refinement | Subject | | | | std. |
|---|---|---|---|---|---|---|
|  |  | $s_1$ | $s_2$ | $s_3$ | $s_4$ |  |
| GB [21] | 0.8715 | 0.9077 | 0.8795 | 0.8943 | 0.8946 | 0.0115 |
| IT [26] | 0.8077 | 0.8797 | 0.8775 | 0.8858 | 0.8733 | 0.0052 |
| CA [18] | 0.7833 | 0.9132 | 0.8870 | 0.9056 | 0.8927 | 0.0119 |
| BT [4] | 0.8782 | 0.9192 | 0.9109 | 0.9102 | 0.9077 | 0.0050 |
| AIM [9] | 0.7393 | 0.9016 | 0.8671 | 0.8945 | 0.8856 | 0.0149 |
| SER [41] | 0.7352 | 0.8801 | 0.8727 | 0.8794 | 0.8852 | 0.0051 |
| SR [23] | 0.7442 | 0.8748 | 0.8855 | 0.8863 | 0.8676 | 0.0090 |
| AWS [17] | 0.7955 | 0.8833 | 0.8817 | 0.8813 | 0.8705 | 0.0059 |
| HFT [32] | 0.7976 | 0.8740 | 0.8803 | 0.8800 | 0.8680 | 0.0058 |
| Judd [27] | 0.8010 | 0.9263 | 0.8804 | 0.9001 | 0.9052 | 0.0189 |
| LG [5] | 0.7790 | 0.9156 | 0.8885 | 0.9109 | 0.9034 | 0.0119 |
| QDCT [40] | 0.7840 | 0.8593 | 0.8556 | 0.8521 | 0.8437 | 0.0067 |
| BMS [50] | 0.8258 | 0.8996 | 0.8657 | 0.8866 | 0.8811 | 0.0140 |

TABLE I: **Saliency performance variation according to subjects.** The AUC values and its standard deviation (*std.*) are reported for each subject and saliency method.

## V. APPLICATIONS

Many applications in personal photo collections have used detection algorithms, attribute learning approaches and various image processing algorithms to construct tag suggestion systems, image search, photo effects, *etc.*. By providing personal attention maps for personal photos, these applications can better represent individual preferences. In this section, we demonstrate three computer vision applications: personal saliency generation, tag region suggestion, and salient region based photo retrieval. The last two algorithms are built on top of the personal saliency generation application. All the experiment settings and dataset used here are consistent with the setting described in Sec. IV. We include the rest of the results partially not presented here in the appendix.

**Personal saliency** In our framework, user attention map generated by our method can be used as another saliency map that is specific to the user only. We first evaluate the performance of saliency estimation according to baseline saliency methods[6]. We report AUC [27] values, which is widely used to assess saliency estimation, in order to demonstrate the performance of our framework with 13 state-of-the-art methods shown in Fig. 9. To understand the individual viewing behaviors, we indirectly observe it by observing independent subject results as shown in Table II. The qualitative results corresponding to Table II is presented in the appendix.

**Region suggestion for tagging** We demonstrate a tag region suggestion application that utilizes photo contents as well as personal saliency maps that suggests preferred candidate regions. We segment the photo via over-segmentation technique [13] and assign segments that overlap with the binarized saliency map as blobs and present the bounding box (BB) as region suggestion.

[6]In this experiment, we conduct tests by replacing the saliency estimation module used in both our implicit calibration and mode seeking steps with each state-of-the-art method.

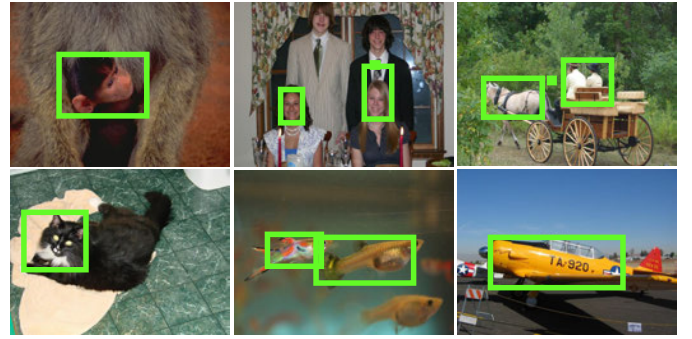

Fig. 10: **Region suggestion for tagging.** With personal saliencies of a photo, our algorithm can provide tag candidate region suggestions.

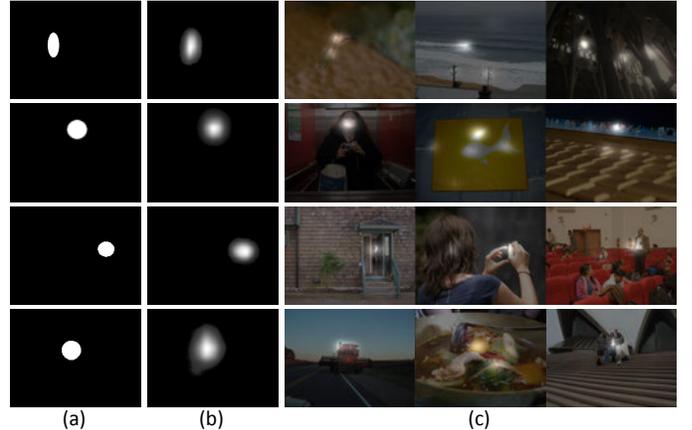

Fig. 11: **Photo retrieval by saliency.** We query regions via (a) mouse drags or (b) human attention maps in order to retrieve photos (c) (displayed by overlaying saliency maps on photos) from vast personal photo collections. When finding a specific photo in a vast collection is difficult, the user can query his or her intended attention region to obtain desired photo.

The results of tag region suggestion are shown in Fig. 10 in which the suggestion is made based on personal saliency information. We assess our results by the PASCAL measure [15] which is defined as $a_o = \frac{area(BB_{dt} \cap BB_{gt})}{area(BB_{dt} \cup BB_{gt})}$, where $BB_{dt}$ and $BB_{gt}$ denotes the estimated and ground truth BBs respectively. By comparing with every ground truth BB, we assign the largest PASCAL measure to each estimated BB. Among multiple candidate regions, by taking the median PASCAL measure for each image, we obtain an average measure of 0.5314. This signifies that roughly more than half of all candidate regions exceed 50% overlap by the PASCAL measure on average. The specific experimental configurations and procedures of this application can be found in the appendix.

While existing tag candidate region suggestions (*e.g.*, Facebook) are made mostly via face detection, our implementation enables tag candidate region suggestions on high attention regions. Thus, tags can be made on the user's interest region including regions which would normally not be suggested as tags by traditional methods.

**Photo retrieval by saliency** Another application we demonstrate is photo retrieval by personal interest location query. When a user possesses vast photo collections and wants to find a specific photo, it may be difficult for a user to search among the vast number of photos or retrieve it by text or color. Since the user is likely to remember where he or she has looked at in the photo, it is possible for the user to query by personal attention regions. The implementation was done by displaying photos with top similarity values [27] compared with the input query. The similarity metric[7] is defined as $Sim(\overline{s}, \overline{t}) = \int \min(\overline{s}(x), \overline{t}(x))\, dx$, for the normalized saliency maps $\overline{s}$ and $\overline{t}$. Fig. 11 shows sampled results of photo retrieval by personal attention queries. More qualitative results with top rank images are shown in the appendix. The user can simply locate interest regions to search photos by a mouse interface. Also, it would be possible to query by gazing at the interest location as an alternate way to query.

## VI. CONCLUSION

In this paper, we present human attention estimation on the fly via implicit gaze estimation and integration with computational saliency through mode seeking. Without any special hardware, our method can estimate the user's attention region on natural scene images. On top of this, users do not need to follow explicit interaction instructions, due to the entirely implicit process. For demonstration of usability and practicality, we introduce three user-specific applications for personal photo collections. For future work, we wish to develop further discussion and applications on human attention.

## APPENDIX

Here, we provide the technical details for efficient incremental GMM clustering and additional application results. In addition, we provide a larger table containing experimental results of the attention map quality using the similarity (Sim.) metric values. We also provide some failure cases seen from our applications and analyze them.

### A. Eye Feature Vector Clustering

The eye feature vector clustering procedure is summarized in Alg. 3. Gaussian mixture model (GMM) is parameterized by $\{\hat{D}_c, \hat{\pi}_c, \hat{\boldsymbol{\mu}}_c, \hat{\boldsymbol{\Sigma}}_c\}_{c=1}^K$, where $K$ is the number of unimodal Gaussian distributions (or clusters), $\{\hat{\boldsymbol{\mu}}_c, \hat{\boldsymbol{\Sigma}}_c\}$ denote the mean and covariance of GMM, $\hat{\pi}_c$ denotes the weights of GMM, and $\hat{D}_c = \sum_{i=1}^{\hat{N}} P(z_i = c|\hat{\mathbf{e}}_i)$. While $\{\hat{D}_c\}$ is not regarded as a model specific parameter traditionally, we explicitly store $\{\hat{D}_c\}$ to reduce the computation complexity at the incremental update stage (used in Line 13 to 15 in Alg. 3). We measure the changes of $\{\boldsymbol{\mu}_c\}$ between subsequent inner iterations to measure the convergence. If the changes are below a threshold $\varepsilon = 0.01$, then we stop the clustering. Also, we fix the maximum number of iteration to 100.

---

[7]Instead of taking the integral, actual computation is done by summation in a discrete manner.

During the training stage, the overall learning step including the batch clustering (Line 1 to 3 in Alg. 3) is conducted on background. After the initial model is constructed, with a time interval, our model is updated by starting from the incremental clustering (Line 5 to 16 in Alg. 3), and then saliency aggregation followed by learning GPR.

**Algorithm 3** Eye feature vector clustering by GMM (incremental GMM).

1: // Initial stage.
2:   Gather $\hat{N}$ number of eye feature vectors $\{\hat{\mathbf{e}}_i\}$.
3:   Batch GMM [2] to produce $\{\hat{D}_c, \hat{\pi}_c, \hat{\boldsymbol{\mu}}_c, \hat{\boldsymbol{\Sigma}}_c\}_{c=1}^K$.
4:
5: // Incremental update.
6:   Gather $N$ number of newly observed eye feature vectors $\{\mathbf{e}_i\}$.
7:   Initialize $\{D_c, \pi_c, \boldsymbol{\mu}_c, \boldsymbol{\Sigma}_c\}_{c=1}^K = \{\hat{D}_c, \hat{\pi}_c, \hat{\boldsymbol{\mu}}_c, \hat{\boldsymbol{\Sigma}}_c\}_{c=1}^K$.
8: **repeat**
9:   ⟨ E-step ⟩
10:    $p_{c,i} \triangleq P(z_i = c|\mathbf{e}_i) = \frac{\pi_c \mathcal{N}(\mathbf{e}_i|\boldsymbol{\mu}_c, \boldsymbol{\Sigma}_c)}{\sum_{c=1}^K \pi_c \mathcal{N}(\mathbf{e}_i|\boldsymbol{\mu}_c, \boldsymbol{\Sigma}_c)}$.
11:   ⟨ M-step ⟩
12:    $D_c = \sum_{i=1}^N p_{c,i}$.
13:    $\pi_c = \frac{\hat{D}_c + D_c}{\hat{N} + N}$.
14:    $\boldsymbol{\mu}_c = \frac{\hat{D}_c \hat{\boldsymbol{\mu}}_c + \sum_{i=1}^N p_{c,i} \mathbf{e}_i}{\hat{D}_c + D_c}$.
15:    $\boldsymbol{\Sigma}_c = \left( \hat{D}_c (\hat{\boldsymbol{\Sigma}}_c + (\hat{\boldsymbol{\mu}}_c - \boldsymbol{\mu}_c)(\hat{\boldsymbol{\mu}}_c - \boldsymbol{\mu}_c)^\top) + \right.$
     $\left. \sum_{i=1}^N p_{c,i} (\mathbf{e}_i - \boldsymbol{\mu}_c)(\mathbf{e}_i - \boldsymbol{\mu}_c)^\top \right) / (\hat{D}_c + D_c)$.
16: **until** converged.

### B. Additional Application Results

Additional results on the applications are shown in the following materials. We show more examples on the personal saliency application and display comparison with state-of-the-art saliency models. The region suggestion for tagging application was conducted by taking images containing somewhat distinguishable objects. The ground truth objectness was recorded by a human subject and the evaluation was done by taking this as the ground truth. Analysis on failure cases are reviewed as well. Lastly, we also show additional results on retrieval by saliency query application.

## C. Personal Saliency

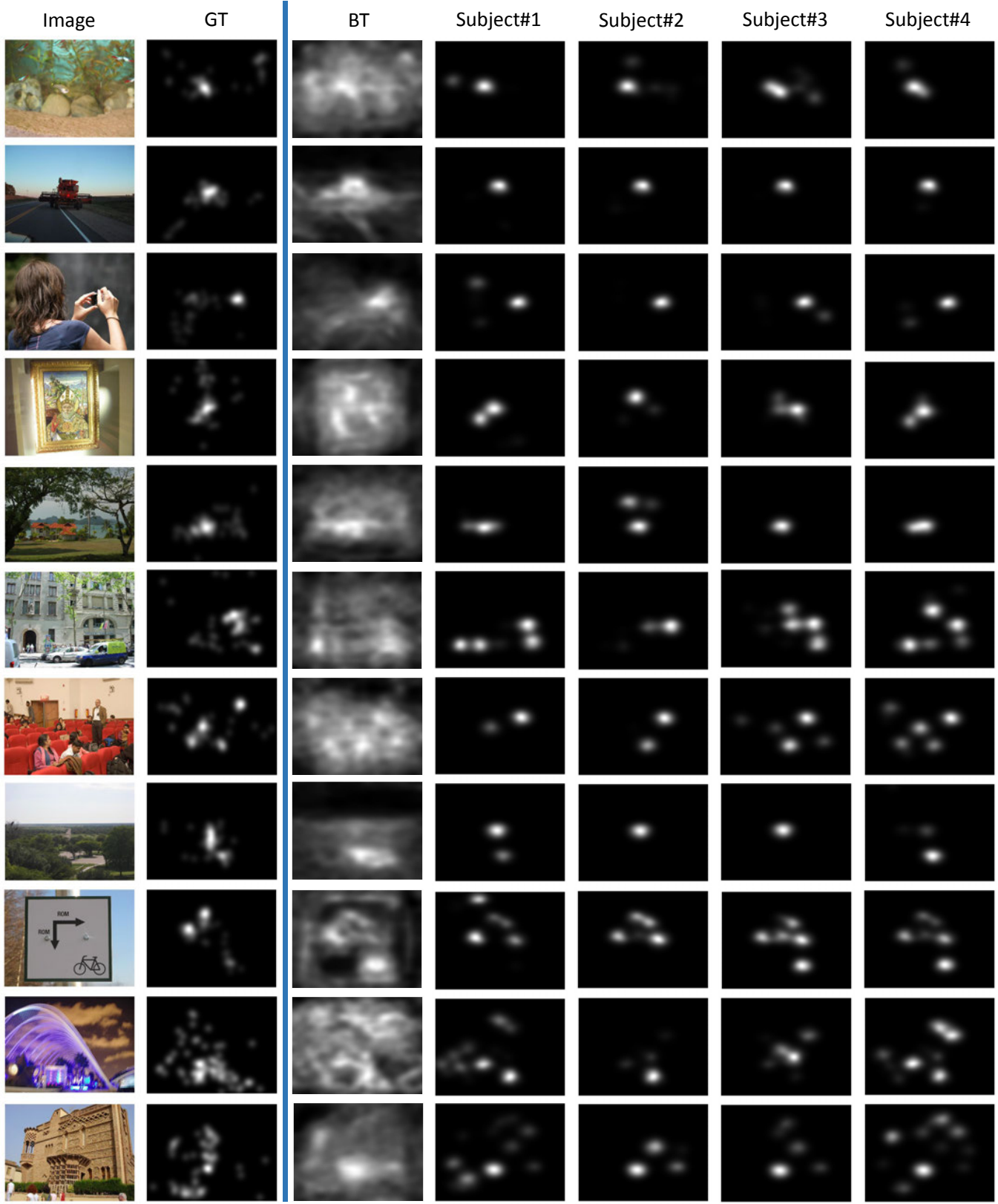

Fig. 12: **Personal saliencies.** From left to right are: input images, ground truth saliency maps, BT [4], refined saliencies from subjects using the baseline BT model. Saliency maps for individual people are shown along with the saliency maps from the BT model, which performed the best among baseline methods.

|     | AUC |     |     |     |     |     |     | Sim. |     |     |     |     |     |     |
| --- | --- | --- | --- | --- | --- | --- | --- | --- | --- | --- | --- | --- | --- | --- |
|     | Baseline | Subject | | | | Subject | | Baseline | Subject | | | | Subject | |
|     |     | $s_1$ | $s_2$ | $s_3$ | $s_4$ | mean | std |     | $s_1$ | $s_2$ | $s_3$ | $s_4$ | mean | std |
| GB  | 0.8715 | 0.9077 | 0.8795 | 0.8943 | 0.8946 | **0.8940** | 0.0115 | **0.3649** | 0.3770 | 0.2525 | 0.3503 | 0.3968 | 0.3442 | 0.0640 |
| IT  | 0.8077 | 0.8797 | 0.8775 | 0.8858 | 0.8733 | **0.8791** | 0.0052 | **0.3284** | 0.3301 | 0.2543 | 0.3120 | 0.3544 | 0.3127 | 0.0426 |
| CA  | 0.7833 | 0.9132 | 0.8870 | 0.9056 | 0.8927 | **0.8996** | 0.0119 | 0.3182 | 0.3979 | 0.2832 | 0.3662 | 0.4194 | **0.3667** | 0.0598 |
| BT  | 0.8782 | 0.9192 | 0.9109 | 0.9102 | 0.9077 | **0.9120** | 0.0050 | 0.3286 | 0.3949 | 0.2989 | 0.3654 | 0.4253 | **0.3711** | 0.0540 |
| AIM | 0.7393 | 0.9016 | 0.8671 | 0.8945 | 0.8856 | **0.8872** | 0.0149 | 0.2724 | 0.3577 | 0.2627 | 0.3511 | 0.3857 | **0.3393** | 0.0532 |
| SER | 0.7352 | 0.8801 | 0.8727 | 0.8794 | 0.8852 | **0.8794** | 0.0051 | 0.3028 | 0.3269 | 0.2355 | 0.3259 | 0.3604 | **0.3122** | 0.0536 |
| SR  | 0.7442 | 0.8748 | 0.8855 | 0.8863 | 0.8676 | **0.8786** | 0.0090 | 0.3007 | 0.3175 | 0.2449 | 0.3209 | 0.3611 | **0.3111** | 0.0484 |
| AWS | 0.7955 | 0.8833 | 0.8817 | 0.8813 | 0.8705 | **0.8792** | 0.0059 | 0.3178 | 0.3418 | 0.2610 | 0.3252 | 0.3707 | **0.3263** | 0.0486 |
| HFT | 0.7976 | 0.8740 | 0.8803 | 0.8800 | 0.8680 | **0.8756** | 0.0058 | **0.3225** | 0.3070 | 0.2484 | 0.3075 | 0.3422 | 0.3013 | 0.0389 |
| Judd | 0.8010 | 0.9263 | 0.8804 | 0.9001 | 0.9052 | **0.9030** | 0.0189 | 0.2902 | 0.4171 | 0.2841 | 0.3775 | 0.4514 | **0.3825** | 0.0722 |
| LG  | 0.7790 | 0.9156 | 0.8885 | 0.9109 | 0.9034 | **0.9046** | 0.0119 | 0.2859 | 0.3938 | 0.2822 | 0.3887 | 0.4216 | **0.3716** | 0.0613 |
| QDCT | 0.7840 | 0.8593 | 0.8556 | 0.8521 | 0.8437 | **0.8527** | 0.0067 | **0.3116** | 0.3178 | 0.2399 | 0.3149 | 0.3313 | 0.3010 | 0.0413 |
| BMS | 0.8258 | 0.8996 | 0.8657 | 0.8866 | 0.8811 | **0.8833** | 0.0140 | 0.3300 | 0.3742 | 0.2612 | 0.3400 | 0.3831 | **0.3396** | 0.0555 |

TABLE II: **AUC and similarity metric values.** The AUC and similarity (Sim.) values are shown for each subject. Comparing the baseline model and our corresponding mean performance, we display the higher performance in bold. The similarity values show that people have diverse viewing patterns. On the contrary, AUC values show that people share a certain "commonness."

## D. Region Suggestion for Tagging

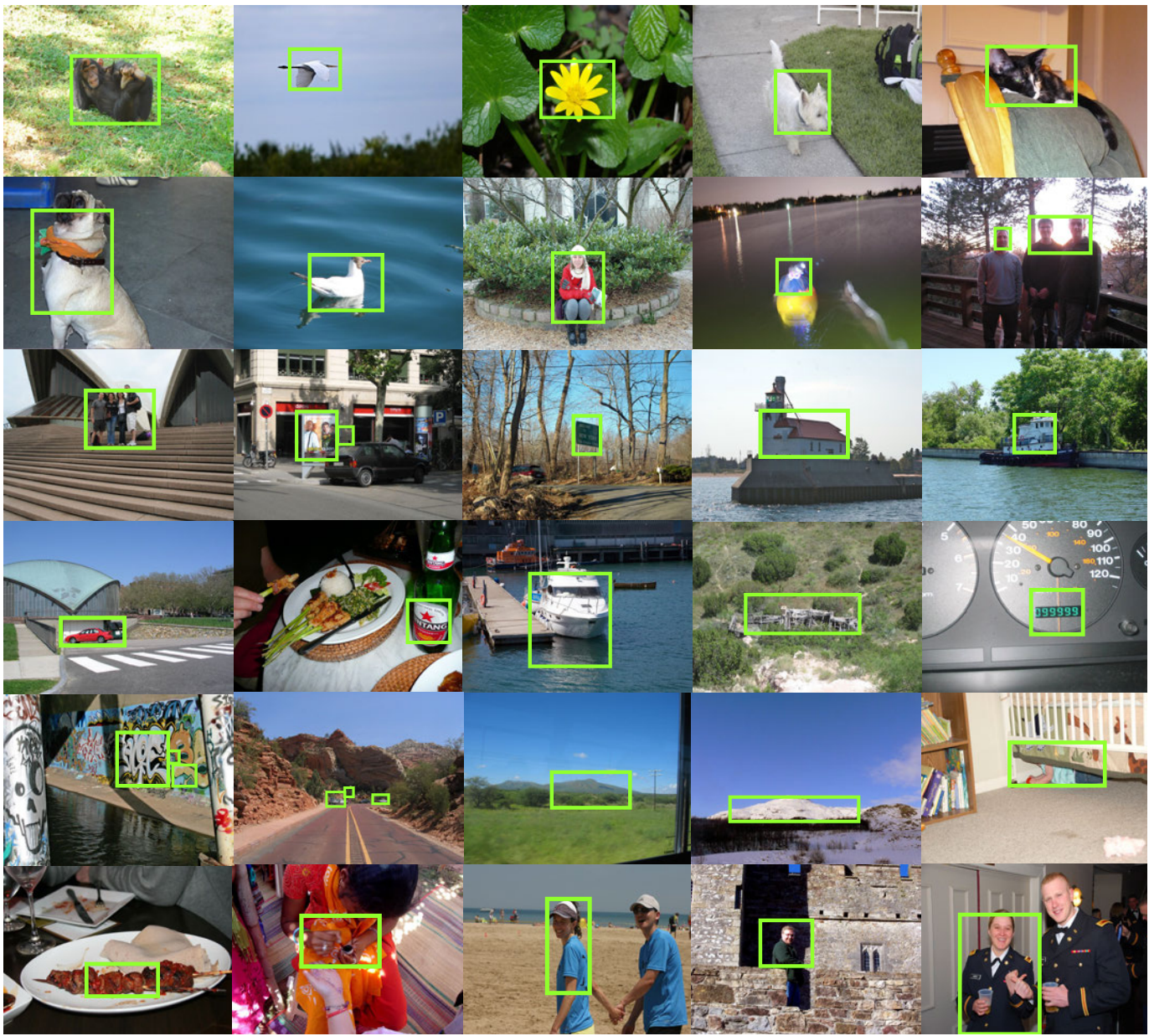

Fig. 13: **Tag region suggestions.** Additional tag suggestion results are shown.

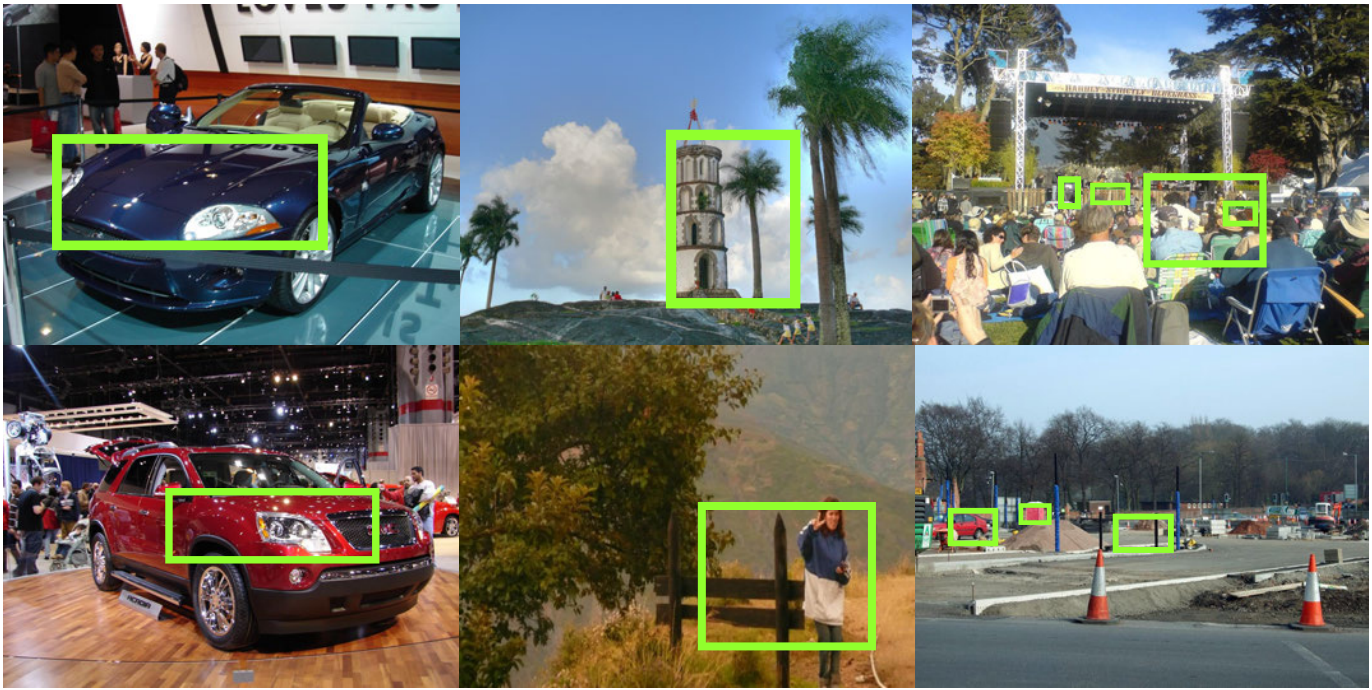

Fig. 14: **Tag region suggestion failure cases.** Some tag region suggestion failures can happen due to [Left] the human nature to look at only partial regions of a large object rather than the object as a whole, [Middle] cases where objects are close together, [Right] complex scenes which involves diverse human attention, etc.

*E. Photo retrieval by saliency*

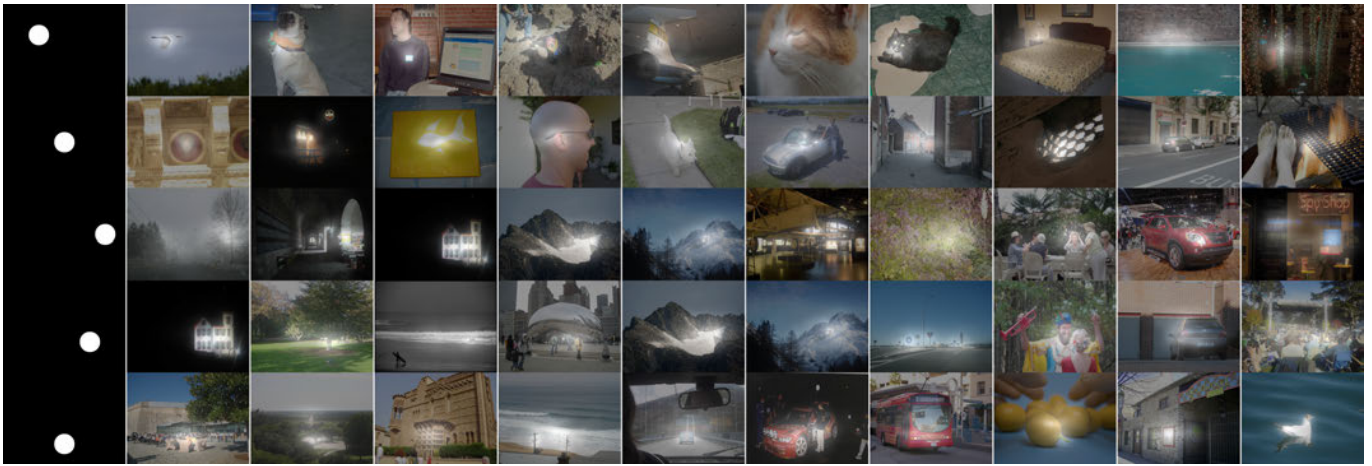

Fig. 15: **Photo retrieval via saliency query.** Additional retrieval results are shown. The bright regions are highlighted via overlaying the attention maps on the images.

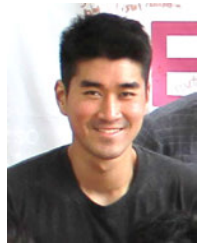
**Jinsoo Choi** received his B.S. and M.S. degrees in Electrical Engineering from Korea Advanced Institute of Science and Technology (KAIST) in 2013 and 2015, respectively. He is currently working towards his Ph.D degree at KAIST. His research interests include multimedia analysis and machine learning. He is a student member of the IEEE.

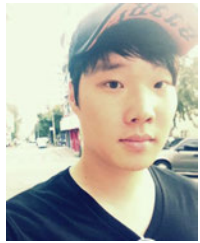
**Tae-Hyun Oh** received the B.E degree (summa cum laude) in Computer Engineering from Kwang-Woon University in 2010, and the M.S degree in Electrical Engineering from KAIST in 2012. He is currently working towards the Ph.D. degree at KAIST, South Korea. He was a visiting student in the Visual Computing Group, Microsoft Research Asia. He was a recipient of Microsoft Research Asia Fellowship, Gold prize of Samsung HumanTech Thesis Award and Qualcomm Innovation Award. His research interests include robust computer vision and machine learning. He is a student member of the IEEE.

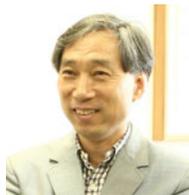
**In So Kweon** received the BS and MS degrees in mechanical design and production engineering from Seoul National University, Seoul, Korea, in 1981 and 1983, respectively, and the PhD degree in robotics from the Robotics Institute, Carnegie Mellon University, Pittsburgh, Pennsylvania, in 1990. He worked for the Toshiba R&D Center, Japan, and joined the Department of Automation and Design Engineering, KAIST, Seoul, Korea, in 1992, where he is now a professor with the Department of Electrical Engineering. He is a recipient of the best student paper runner-up award at the IEEE Conference on Computer Vision and Pattern Recognition (CVPR 09). His research interests are in camera and 3D sensor fusion, color modeling and analysis, visual tracking, and visual SLAM. He was the program co-chair for the Asian Conference on Computer Vision (ACCV 07) and was the general chair for the ACCV 12. He is also on the editorial board of the International Journal of Computer Vision. He is a member of the IEEE and the KROS.